\documentclass[a4paper,fleqn]{cas-dc}

\usepackage[numbers]{natbib}
\usepackage{amssymb}
\usepackage{amsmath,amssymb,amsfonts}
\usepackage{algorithmic}
\usepackage{textcomp}
\usepackage{xcolor}
\usepackage{svg}
\usepackage{ulem}
\usepackage{amsmath}
\usepackage[export]{adjustbox}
\usepackage[singlelinecheck=false]{caption}

\usepackage{caption}
\usepackage{subcaption}

\def\tsc#1{\csdef{#1}{\textsc{\lowercase{#1}}\xspace}}
\tsc{WGM}
\tsc{QE}
\tsc{EP}
\tsc{PMS}
\tsc{BEC}
\tsc{DE}
\begin{document}
\let\WriteBookmarks\relax
\def\floatpagepagefraction{1}
\def\textpagefraction{.001}

% Short title
\shorttitle{Qualitative Data Augmentation for Performance Prediction in VLSI circuits}

% Short author
\shortauthors{et al.}

% Main title of the paper
\title [mode = title]{Qualitative Data Augmentation for Performance Prediction in VLSI Circuits}%{This is a specimen $a_b$ title}                      
% Title footnote mark
% eg: \tnotemark[1]
% \tnotemark[1,2]

% Title footnote 1.
% eg: \tnotetext[1]{Title footnote text}
% \tnotetext[<tnote number>]{<tnote text>} 
% \tnotetext[1]{This document is the results of the research
%   project funded by the National Science Foundation.}

% \tnotetext[2]{The second title footnote which is a longer text matter
%   to fill through the whole text width and overflow into
%   another line in the footnotes area of the first page.}

% First author
%
% Options: Use if required
% eg: \author[1,3]{Author Name}[type=editor,
%       style=chinese,
%       auid=000,
%       bioid=1,
%       prefix=Sir,
%       orcid=0000-0000-0000-0000,
%       facebook=<facebook id>,
%       twitter=<twitter id>,
%       linkedin=<linkedin id>,
%       gplus=<gplus id>]
% \author{\IEEEauthorblockN{1\textsuperscript{st} Prasha Srivastava}
% \IEEEauthorblockA{\textit{CVEST and CSTAR} \\
% \textit{IIIT, Hyderabad}\\
% Hyderabad, India \\
% prasha.srivastava@research.iiit.ac.in}
\author[1]{Prasha Srivastava}
% [type=editor,
%                         auid=000,bioid=1,
%                         prefix=Sir,
%                         role=Researcher,
%                         orcid=0000-0001-7511-2910]

% Corresponding author indication
\cormark[1]

% Footnote of the first author
% \fnmark[1]

% Email id of the first author
\ead{prasha.srivastava@research.iiit.ac.in}%cvr_1@tug.org.in}

% URL of the first author
% \ead[url]{www.cvr.cc, cvr@sayahna.org}

%  Credit authorship
% \credit{Conceptualization of this study, Methodology, Software}

% Address/affiliation
% \affiliation[1]{organization={Elsevier B.V.},
%     addressline={Radarweg 29}, 
%     city={Amsterdam},
%     % citysep={}, % Uncomment if no comma needed between city and postcode
%     postcode={1043 NX}, 
%     % state={},
%     country={The Netherlands}}
\affiliation[1]{organization={International Institute of Information Technology},
    % addressline={Hyderabad}, 
    city={Hyderabad},
    % citysep={}, % Uncomment if no comma needed between city and postcode
    postcode={500032}, 
    % state={},
    country={India}}

% Second author
\author[1]{Pawan Kumar}%[style=chinese]
\ead{pawan.kumar@iiit.ac.in}%cvr_1@tug.org.in}
% \fnmark[1]
% Third author
\author[1]{Zia Abbas}%[%role=,suffix=,]
\ead{zia.abbas@iiit.ac.in}%cvr_1@tug.org.in}
% \fnmark[1]
% \ead{cvr3@sayahna.org}
% \ead[URL]{www.sayahna.org}

% \credit{Data curation, Writing - Original draft preparation}

% Address/affiliation
% \affiliation[2]{organization={Sayahna Foundation},
%     % addressline={}, 
%     city={Jagathy},
%     % citysep={}, % Uncomment if no comma needed between city and postcode
%     postcode={695014}, 
%     state={Trivandrum},
%     country={India}}

% Fourth author
% \author%
% [1,3]
% {author4}
% \cormark[2]
% \fnmark[1,3]
% \ead{rishi@stmdocs.in}
% \ead[URL]{www.stmdocs.in}

% \affiliation[3]{organization={STM Document Engineering Pvt Ltd.},
%     addressline={Mepukada}, 
%     city={Malayinkil},
%     % citysep={}, % Uncomment if no comma needed between city and postcode
%     postcode={695571}, 
%     state={Trivandrum},
%     country={India}}

% Corresponding author text
\cortext[cor1]{Corresponding author}
% \cortext[cor2]{Principal corresponding author}

% Footnote text
% \fntext[fn1]{This is the first author footnote. but is common to third
%   author as well.}
% \fntext[fn2]{Another author footnote, this is a very long footnote and
%   it should be a really long footnote. But this footnote is not yet
%   sufficiently long enough to make two lines of footnote text.}

% For a title note without a number/mark
% \nonumnote{This note has no numbers. In this work we demonstrate $a_b$
%   the formation Y\_1 of a new type of polariton on the interface
%   between a cuprous oxide slab and a polystyrene micro-sphere placed
%   on the slab.
%   }

% Here goes the abstract
\begin{abstract}
Various studies have shown the advantages of using Machine Learning (ML) techniques for analog and digital IC design automation and optimization. Data scarcity is still an issue for electronic designs, while training highly accurate ML models. This work proposes generating and evaluating artificial data using generative adversarial networks (GANs) for circuit data to aid and improve the accuracy of ML models trained with a small training data set. The training data is obtained by various simulations in the Cadence Virtuoso, HSPICE, and Microcap design environment with TSMC 180nm and 22nm CMOS technology nodes. The artificial data is generated and tested for an appropriate set of analog and digital circuits. The experimental results show that the proposed artificial data generation significantly improves ML models and reduces the percentage error by more than 50\% of the original percentage error, which were previously trained with insufficient data. Furthermore, this research aims to contribute to the extensive application of AI/ML in the field of VLSI design and technology by relieving the training data availability-related challenges.
% This template helps you to create a properly formatted \LaTeX\ manuscript.
% \noindent\texttt{\textbackslash begin{abstract}} \dots 
% \texttt{\textbackslash end{abstract}} and
% \verb+\begin{keyword}+ \verb+...+ \verb+\end{keyword}+ 
% which
% contain the abstract and keywords respectively. 

% \noindent Each keyword shall be separated by a \verb+\sep+ command.
\end{abstract}

% Use if graphical abstract is present
% \begin{graphicalabstract}
% \includegraphics{figs/grabs.pdf}
% \end{graphicalabstract}

% Research highlights
% \begin{highlights}
% \item Research highlights item 1
% \item Research highlights item 2
% \item Research highlights item 3
% \end{highlights}

% Keywords
% Each keyword is seperated by \sep
\begin{keywords}
Machine Learning \sep Artificial Intelligence \sep VLSI Design \sep Generative Adversarial Networks 
% quadrupole exciton \sep polariton \sep \WGM \sep \BEC
\end{keywords}

\maketitle

% \section{Introduction}
\section{Introduction and Related Work}
\label{sec:sample1}
Machine Learning (ML) is an Artificial Intelligence (AI) method that allows computers to act without requiring definitive programming. ML assists in improving the accuracy of prediction for outcomes in various applications these days. Predictions are made by models using sample data referred to as training data. With the advent of machine learning, the field of VLSI design and testing has the prospect of achieving high levels of automation, speed, and efficiency. In the past, many AI/ML approaches have been introduced to achieve noteable results in the VLSI design domain \cite{b12,b25,b26,b27,b28,b29}.

Most of these approaches are highly dependent on a large amount of training data, which is a challenge for many applications. Other fields, such as sound classification, medical image analysis, disease diagnosis, face recognition, etc., employing AI/ML techniques also face the issue of data scarcity. Data scarcity can be defined as the lack of sufficient quantity or diversity in training data that can increase the learning ability of a machine learning model. As discussed by A. Munappy et al. \cite{b13}, in real-world industrial applications of deep learning in different domains, the shortage of diverse data is one of the challenges that can significantly impact the overall performance of deep learning systems. Various studies have been done to overcome this challenge. Data augmentation and synthetic data generation have been used to overcome this challenge and build large-scale training datasets in these fields. Data augmentation with synthetically created samples has been proven beneficial for several machine learning models \cite{b14}.

\begin{figure}[t!]
\centering
\includegraphics[scale=0.23]{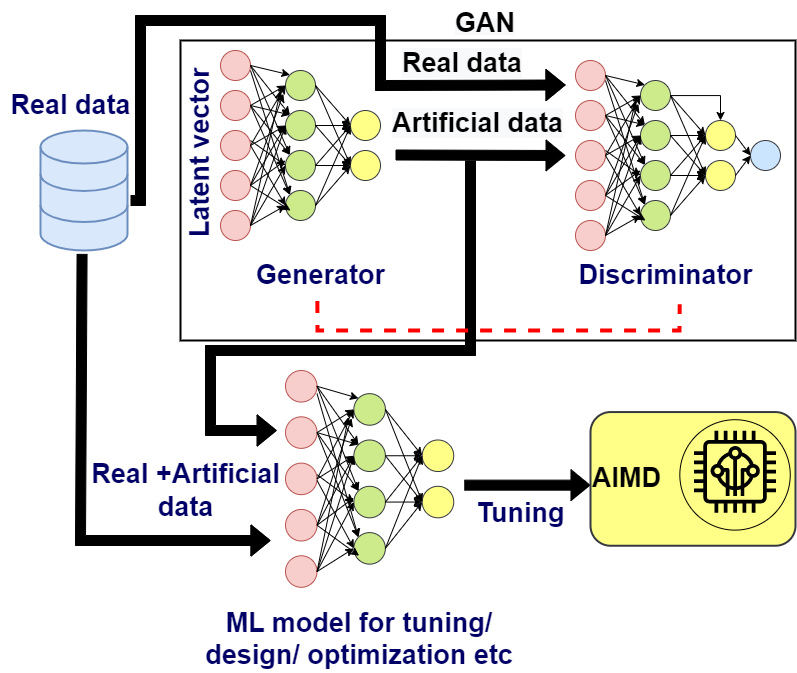}
\caption{\label{figure:1b}An example of machine learning based data augmentation method using GANs for tuning AIMDs}
\end{figure}

A. Mikołajczyk and M. Grochowski \cite{b15} compared and analyzed multiple data augmentation methods in image classification and improved the training process efficiency for image classification. To solve the problem of class imbalance due to the lack of fraudulent electricity consumers, M. Asif et al. \cite{b16} proposed employing an evolutionary bidirectional Wasserstein Generative Adversarial Network (Bi-WGAN). They \cite{b16} use Bi-WGAN to synthesize the most plausible fraudulent electricity consumer samples to detect non-technical losses (NTL) in smart meters. W. Tan and H. Guo \cite{b17} also utilized a data augmentation method in their automatic COVID-19 diagnosis framework from lung CT images and improved the generalization capability of the 2D CNN classification models. Their data augmentation method uses multiple Hounsfield Unit (HU) normalization windows. HU window determines the variety of tissue densities visible in the CT images. Kortylewski et al. \cite{b19} used synthetically generated data to reduce the number of real-world images needed for training deep face recognition systems and simultaneously achieved an increased performance.
Synthetic data has also been used for facial expression analysis \cite{b20}, signal generation \cite{b21}, industrial data generation \cite{b22}, etc.

%(Importance of data augmentation in fields,Existing data augmentation work, use of GAN)

For image augmentation, various algorithms such as geometric transformations, color space augmentations, kernel filters, mixing images, random erasing, feature space augmentation, adversarial training, generative adversarial networks, neural style transfer, meta-learning, etc. are used \cite{b18}. Existing work regarding data augmentation and synthetic data generation deals with image data. It uses image-based evaluation measures such as inception score, Frechet inception distance (FID), average log-likelihood, Parzen window estimates, visual fidelity, etc. Generative adversarial networks and variational autoencoders have been used for many image data augmentation tasks \cite{b21,b22,b23}. GAN has also shown promising results for synthesizing artificial data for Intrusion Detection Systems (IDS), medical records, educational records etc \cite{b41,b42}. Generative models are known to produce large and diverse synthetic data for image datasets. Data augmentation has improved ML models for image-related tasks; data augmentation or artificial data generation can potentially enhance the ML models working on circuit data. For electronic circuits, circuit data refers to data regarding component values, temperature, process, output variations, node voltages, branch currents, delays, power, etc. Hence, a synthetic data generation method for augmenting circuit data using generative adversarial networks (GAN) is proposed for the first time in this work. Our contribution is summarized below:
\begin{enumerate}
    \item Data augmentation for electronic circuit data. 
    \item Prevention of mode collapse. Mode collapse is observed when the generated artificial data is repetitive and covers only few recurring samples.
    \item Reduction in prediction error of a previously trained ML model using the generated artificial data for delay prediction in complex circuits. Additionally, the artificial data for basic circuits is scalable to complex and larger circuits.
\end{enumerate}
This paper is organized as follows. Section 2 presents a detailed description of the used circuit data. This section also answers why data augmentation is needed for circuit data and various applications where data augmentation for circuit data can prove beneficial are discussed. Section 3 briefly introduces GAN architecture, working, and issues such as mode collapse. Section 3 further presents ways to prevent mode collapse. Section 4 offers the setup of experiments and performance evaluation methodology followed by the obtained experimental results. Finally, in Section 5, the conclusions are drawn. 
\begin{table*}
\centering 
\begin{tabular}{|p{5cm}|p{10cm}|p{1.3cm}|}
\hline 
    Dataset  & Applications & \#Features   \\ \hline 
    Current Reference Circuit   & Building block for oscillators, amplifiers, and phase-locked loops. Used in AIMDs such as pacemakers, communication devices, etc. \cite{b33} & 9   \\ \hline 
    Low Dropout regulator (LDO)   &  In wired and wireless communications for portable battery-powered equipment, implanted biomedical devices, automotive applications, digital Core supply, and consumer electronics. \cite{b34,b35,b36} & 12 \\ \hline 
    Operational Trans-conductance Amplifier  &  For designing basic voltage amplifiers, active filters, etc. For example, it finds applications in  biomedical instruments. \cite{b33} & 7   \\ \hline 
    Comparator Circuit  &  For interfacing with digital logic in electronic devices \cite{b37,b38} & 5   \\ \hline 
    Voltage Reference &  For designing power supplies, measurement and control systems, DACs, and ADCs. These are used in high-precision applications like medical and scientific equipment. \cite{b33} & 6   \\ \hline 
    Temperature Sensor & In medical devices, handling chemicals, food processing, etc.  \cite{b39,b40} & 6  \\ \hline 
\end{tabular}
\caption{\label{table:data}Different analog circuits with their practical applications which we have used in this work.}
\end{table*}
\section{Description of Circuit Data and Need for Data Augmentation}
The hardware for all the applications in healthcare, mobility, the internet of things, wearable and implantable devices, etc., consists of analog and digital electronic circuitry. Data extracted from these circuits is helpful if used to train a machine learning model. For example, we can predict the working condition of the device from data, find the faulty component within the device, use this data for tuning purposes and correct any minor faults, or use data for designing circuits with the help of a proper machine learning model \cite{b27,b28}. The necessary circuit data for these purposes will mainly consist of important circuit parameters and the actual output of the circuit.

Getting sufficient circuit data is an important issue, especially when the circuit is deployed already or is unreachable or extracting the data is limited by expense, resources, and privacy issues. For example, once implanted within a patient's body, biomedical devices such as active implantable medical devices (AIMDs) cannot be tuned manually, but can be fine tuned by programming with new operational parameters \cite{b43}. ML can aid this tuning process. But the data required is limited by patient privacy and other issues. Data augmentation solves this insufficient data issue and helps train and develop a good ML model for such tasks (refer Figure \ref{figure:1b}). Another example where data augmentation can be helpful is when it is computationally expensive to obtain a large amount of training data for tasks such as design automation, optimization, testing, etc.  

\subsection{Description of Circuit Data}
%%\sout{\textcolor{red}{Describe circuit data and various applications, include a table.}}

We have used six datasets from six widely used analog electronic circuits and fourteen delay datasets from fourteen basic digital circuits for this work. Datasets were collected using EDA tools Cadence Virtuoso \cite{b30}, Micro-Cap \cite{b31}, and HSpice \cite{b32}. The training data was generated by varying process variables (process, temperature, and supply) and design variables (transistors widths/lengths, capacitor and resistor values, etc) for a given circuit topology.
%Figure \ref{figure:55} shows the training data generation process.
The practical applications of the analog circuits are summarized in Table \ref{table:data}. For digital circuits, refer Table \ref{table:digitaldata}. The data for each analog circuit consists of parameters such as output current, MOSFET process corners, temperature, supply voltage, resistance, etc. Similarly for digital circuits, it consists of parameters such as temperature, supply voltage, MOSFET width, MOSFET length, oxide thickness, delay for output nodes wrt. different input nodes etc.

We take the circuit parameters, which are most likely to vary while the device is working, or the parameters that can affect the functionality of the circuit; for example, leakage currents and corner variations. This data can be helpful for many tuning and troubleshooting purposes \cite{b12}.

\tiny
% \begin{table*}
% \centering 
% \begin{tabular}{|p{4cm}|p{12cm}|}
% \hline 
%     Dataset  & Applications   \\ \hline 
%     Current reference circuit   & Building block for oscillators, amplifiers, and phase-locked loops. Used in AIMDs such as pacemakers, communication devices, etc.    \\ \hline 
%     Low dropout regulator (LDO)   &  In wired and wireless communications for portable battery-powered equipment, automotive applications, digital Core supply, and consumer electronics. \\ \hline 
%     Operational Trans-conductance Amplifier  &  For designing basic voltage amplifiers, active filters, etc. For example, it finds applications in  biomedical instruments.    \\ \hline 
%     Comparator circuit  &  For interfacing with digital logic in electronic devices   \\ \hline 
%     Voltage Reference &  For designing power supplies, measurement and control systems, DACs, and ADCs. These are used in high-precision applications like medical and scientific equipment.   \\ \hline 
%     Temperature Sensor & In medical devices, handling chemicals, food processing, etc.    \\ \hline 
% \end{tabular}
% \caption{\label{table:data}Different circuits with their practical applications which we have used in this work.}
% \end{table*}

\normalsize 

\begin{table}
\centering 
% \begin{tabular}{|c|c|c|}
\begin{tabular}{|p{6.0cm}|p{1.2cm}|}
\hline 
    Dataset  & \#Features \\ \hline 
    NOT gate delay   & 17 \\ \hline
    Two input NAND gate delay   & 19 \\ \hline 
    Two input AND gate delay    & 19 \\ \hline 
    Two input NOR gate delay    & 19 \\ \hline
    Two input OR gate delay    & 19 \\ \hline
    Two input XOR gate delay    & 19 \\ \hline
    Three input AND-OR circuit delay    & 21 \\ \hline
    Full adder delay    & 21 \\ \hline
    2:1 Multiplexer delay    & 21 \\ \hline
    Three input NAND gate delay    & 21 \\ \hline
    Three input AND gate delay    & 21 \\ \hline
    Three input NOR gate delay    & 21 \\ \hline 
    Four input AND-OR circuit (AO22) delay   & 23 \\ \hline
    Four input AND-OR circuit (AO31) delay   & 23 \\ \hline
    
\end{tabular}
\caption{\label{table:digitaldata}List of digital circuit datasets used in this work.}
\end{table}

% \begin{figure}%
%     \centering
%     %\subfloat[\centering Current reference circuit]{{\includegraphics[max size={\columnwidth	}{\textheight}]{Untitled Diagram-Page-3.drawio.png} }}%
%     \subfloat[\centering LDO circuit]{{\includegraphics[max size={\columnwidth	}{\textheight}]{LDO2.drawio (2).png} }}
%     \qquad
%     %\subfloat[\centering LDO circuit]{{\includegraphics[max size={\columnwidth	}{\textheight}]{LDO1.drawio (3).png} }}%
%     \subfloat[\centering Current reference circuit]{{\includegraphics[max size={\columnwidth	}{\textheight}]{Current_ref1.drawio.png} }}
%     \caption{Circuits used....}%
%     \label{fig:example}%
% \end{figure}

% \begin{figure}[htp] 
%     \centering
%     %\subfloat[data a]{%
%         \includesvg[width=0.4\columnwidth]{Current_ref1.svg}%{Current_ref1_drawio.svg}%
%       %
%         \caption{ \label{figure:1}Circuits}
%      %   }%
%     %\hfill%
% \end{figure}

% \begin{figure}[htp]
%     %\subfloat[data b]{%
%         \includesvg[width=\columnwidth]{LDO2.svg}%{LDO2_drawio.svg}%
%         %
%      %   }%
%     \caption{\label{figure:2}Circuits}
% \end{figure}

\subsection{Need of Data Augmentation for Circuit Data}
%%\sout{\textcolor{red}{Describe why data augmentation is crucial for embedded devices.}}

These days various devices are being heavily used in applications where they are expected to stay and function for an extended period of time. It may be possible that these devices are inconvenient to reach once they are in use. In case of any malfunction, the electronic components in these devices must be tuned to achieve proper functioning. A machine learning model can carry out this tuning process externally, but massive data is needed to train the ML model to achieve perfect functioning. An example of such device is AIMDs. Still, AIMDs include pacemakers, ventricular assist devices, deep brain stimulators, implantable hearing aids, etc. Once implanted inside a user's body, these devices are inconvenient to reach.

Gathering this massive data from its electronic components and circuits is a tremendous challenge. This data collection process is currently limited by concerns such as:
\begin{enumerate}

\item Data may have privacy issues or data may be proprietary.
\item It may be computationally expensive or power consuming to obtain data.
\item Data may be practically difficult to obtain.

\end{enumerate}
Moreover, many ML applications in VLSI design deal with automation in designing and testing circuits. ML is also proposed for design optimization tasks. Data is the primary requirement for training good and accurate machine learning models for the above-mentioned tasks. Data augmentation can be helpful when it is computationally expensive or time-consuming to obtain a large amount of training data.

Hence, data augmentation and synthetic data generation can prove helpful for training accurate ML models for such applications. 
% \begin{figure}[h!]
%     \includegraphics[max size={\columnwidth	}{\textheight}]{Untitled Diagram-Page-1.drawio.png}
%     \caption{\label{figure:3}Use of GAN for AIMDs}
%     \end{figure}
% \begin{figure}[h!]
% \centering 
%     \includegraphics[max size={0.7\columnwidth	}{\textheight}]{Tuning using ML model.drawio.png}
%     \caption{\label{figure:1a}Use of Machine learning for tuning AIMDS}
%     \end{figure}
    
% \begin{figure}[h!]
% \centering 
%     \includegraphics[max size={0.7\columnwidth	}{\textheight}]{Tuning using GAN.drawio.png}
%     \caption{\label{figure:1b}Use of GANs for using Machine learning for tuning AIMDs}
%     \end{figure}

% \begin{figure}[h]
% \centering 
%     % \includegraphics[scale=0.3]{Tuning using ML model.drawio.png}
%     \includegraphics[scale=0.425]{Tuning using ML model.drawiov2.png}
%     \caption{\label{figure:1b}Use of Machine learning for tuning AIMDs.}
% \end{figure}

\section{Data Augmentation Using Generative Adversarial Networks(GANs)}
\subsection{Description of GAN Architecture}
GANs are a type of generative machine learning model that tries to learn the data distribution and create synthetic data.
GAN architecture consists of two deep neural networks, one of which we call a generator, and the other is called a discriminator. The training data, which is the actual existing data, is referred to as real data, and the data generated by GAN is referred to as artificial data here. 
 
The generator has the task of creating the samples intended to come from the same distribution as the real data. The generator takes a random noise vector as input. This vector represents the latent features of the data generated. For example, this vector represents features like shape and color for image data. The generator gives artificial samples at the output.

\begin{figure}[t!]
\centering
\includegraphics[scale=0.32]{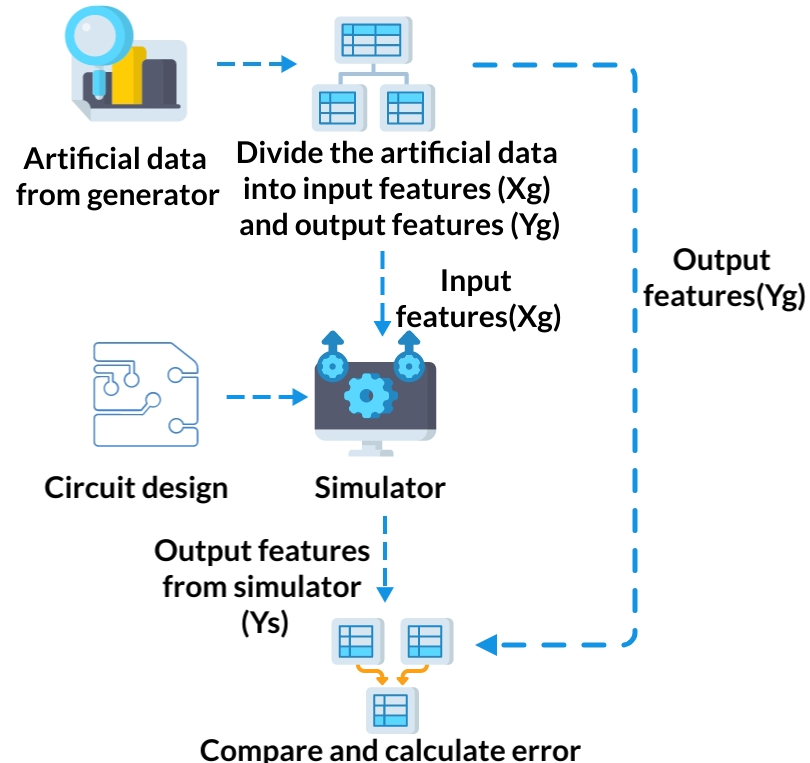}
\caption{\label{figure:56}Validation process for artificially generated data.}
\end{figure}

\begin{figure*}[t!]
\centering
\includegraphics[scale=0.32]{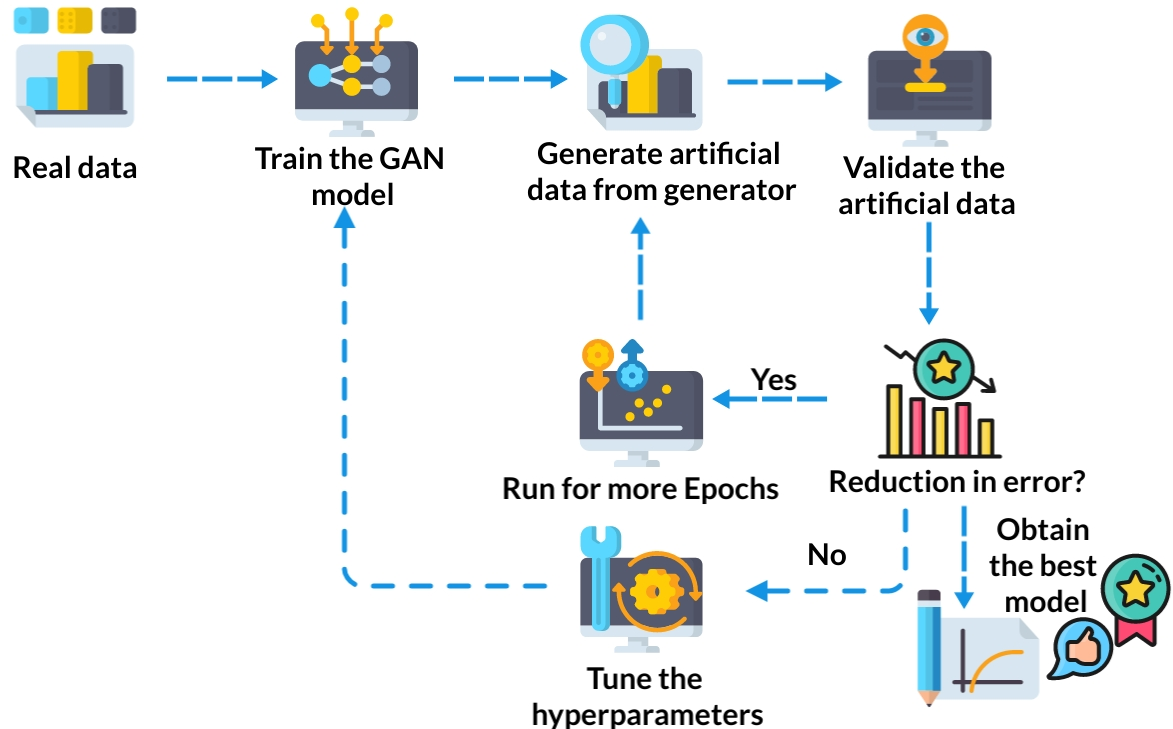}
\caption{\label{figure:57}Complete training process.}
\end{figure*}
 
The discriminator examines the samples from the generator and the real samples and tells if they are real or artificial. Thus, we can say that discriminator is just a conventional classifier that classifies input data into two classes: real or artificial. Discriminator, in this way, learns the features of the real data.

Conceptually discriminator then provides feedback to the generator, which helps the generator create data resembling real data. Technically the generator is trained to develop data towards what the discriminator thinks is real. Both networks are trained alternatively, and they compete to improve themselves. Eventually, the discriminator identifies the tiny difference between the real and the generated, and the generator creates data such that the discriminator cannot differentiate anymore. The GAN is then said to have reached convergence and now can produce data resembling natural data.

\textbf{Loss Function for GAN:} GAN training process can be written as the following optimization objective: 
\begin{equation}
 \min_{G} \max_{D} E_{x\sim X} \log(D(x)) + E_{z\sim Z} \log(1 - D(G(z))). \label{eqn:minmax}
%min_{G}  max_{D}  E_(x∼X)  log(D(x)) + 〖E 〗_(z∼Z) log(1 - D(G(z))) 
\end{equation}

Here $X$ is the original real data, and $Z$ is the latent variable distribution. The above optimization objective translates to $D$, the discriminator wanting to maximize the classification performance between real and artificial samples and G, the generator wanting to minimize the same. The above Equation \eqref{eqn:minmax} is the binary cross entropy loss between real and artificial samples.

Practically, we would want to optimize both $G$ and $D$ separately at each iteration: 
\begin{itemize}
\item Optimization objective for discriminator:
\begin{equation*}
\max_{D} E_{x\sim X} \log(D(x)) + E_{z\sim Z} \log(1 - D(G(z))). 
\end{equation*}
\item Optimization objective for generator:
\begin{equation*}
%(min )¦G 〖E 〗_(z∼Z) log(1 - D(G(z)))
\min_{G} E_{z\sim Z} \log(1 - D(G(z))). 
\end{equation*}
\end{itemize}
But such a minimax game does not perform well in practice because when the discriminator rejects the generator sample with high confidence then \(D(G(z)) = 0\). Thus, the optimization objective for generator %{\(J^G\)} 
vanishes and the gradient for generator vanishes with it. Due to this the generator is now stuck at poor performance. 

\subsection{Preventing Mode collapse in GANs}
%%\sout{\textcolor{red}{Describe spectral normalization and regularization; write in your words, dont copy paste, cite the original papers!}}

Other deep learning models are trained to achieve a single minima, whereas GANs are trained to achieve equilibrium between two networks working as adversaries. The model parameters may oscillate, destabilize, and never converge, making the training unstable. Due to this instability, a common issue faced in GANs is mode collapse \cite{b1,b2,b3}. %%\sout{\textcolor{red}{Use cite command to cite a reference}}where the generator collapses and produces samples of a particular mode while ignoring the other modes.

Different approaches have been proposed to address the problem of mode collapse in GANs. Martin Arjovsky et al. proposed WGAN \cite{b2}, where the objective function for training was changed to Wasserstein distance. Ishaan Gulrajani et al. proposed adding the gradient penalties in the objective function to enforce the Lipschitz constraint to improve the training of WGAN \cite{b3}. Various approaches used multiple generators \cite{b4,b6,b7}. Unrolled GANs \cite{b5} were proposed to prevent the generator from overfitting for a particular discriminator.
Input-based regularizations \cite{b3} had drawbacks while imposing regularization on the space outside of the supports of the generator and data distributions. 

The standard form for GAN is given by Equation \eqref{eqn:minmax}, which can be further written as
 \begin{align*}
     \min_{G} \max_{D} F(G,D),
 \end{align*}
Spectral normalization proposed by Takeru Miyato et al. \cite{b9} has presented more promising results while tackling mode collapse in GANs. They target to find the discriminator $D$ from a set of $K$ Lipschitz continuous functions to stabilize the training of the discriminator
 \begin{equation*}
    \arg \max_{||f||_{lip}<K} F(G,D),
 \end{equation*}
where $||f||_{lip}$ is the Lipschitz constant of the discriminator function $f.$
Their approach constrains the spectral norm of each discriminator layer to control the Lipschitz constant of the overall discriminator function.
Spectral normalization as given by Takeru Miyato et al. \cite{b9} is as follows
 \begin{align*}
     W_{SN}(W):= W/\sigma(W), 
 \end{align*}
where $W$ is the weight matrix of a layer, $\sigma(W)$ is the spectral norm of the matrix $W$, which is equivalent to the largest singular value of $W$. They make sure that the Lipschitz constant of the discriminator function is bounded by normalizing weights of each layer such that $\sigma(W_{SN}(W))=1$.

%%\sout{\textcolor{blue}{WHAT IS SPECTRAL NORMALIZATION}}

Kanglin Liu et al. \cite{b8} proposed spectral regularization to solve the continuing mode collapse issue even in the Spectral Normalized GANs(SN-GANs) \cite{b9}. Spectral regularization is based on the observation that mode collapse and spectral collapse in discriminator's weight matrices go hand in hand; moreover, they demonstrated that spectral collapse is the cause of mode collapse. They define spectral collapse as the vanishing of many singular values of a matrix. Spectral collapse for a discriminator with spectral normalization can be explained as a considerable decrease of singular values of $W_{SN}(W)$ in the discriminator. The weight matrix after applying singular value decomposition can be represented as
\begin{equation*}
     W = U \cdot \Sigma \cdot V^{T}, 
 \end{equation*}
where $U$ and $V$ are orthogonal matrix, and $\Sigma$ is given as 
\begin{equation*}
     \Sigma = \begin{bmatrix}
D & 0 \\
0 & 0 
\end{bmatrix},
 \end{equation*}
where $D$ represents spectral distribution of $W$ as follows
\begin{equation*}
     D = \begin{bmatrix}
    \sigma_{1} & & &\\
    & \sigma_2 & &\\
    & & \ddots & \\
    & & & \sigma_{r}
  \end{bmatrix}.
 \end{equation*} 
 To avoid spectral collapse, the following steps are followed to obtain spectral regularized weights:
 \begin{enumerate}
     \item To compensate $D$, $\Delta D$ is found, where $\Delta D$ is given by
     \begin{equation*}
     \Delta D = \begin{bmatrix}
    {\sigma_{1}-\sigma_{1}} & \hdots & \hdots & \hdots & \hdots & 0\\
    0 & \ddots & \hdots & \hdots & \hdots & \vdots \\
    %& & \ddots & & \\
    \vdots & \hdots & {\sigma_{1}-\sigma_{i}} & \hdots & \hdots & \vdots\\
    \vdots & \hdots & \hdots & 0 & \hdots & \vdots \\
    \vdots & \hdots & \hdots & \hdots & \ddots & \vdots \\
    0 & \hdots & \hdots & \hdots &\hdots & 0 \\ 
  \end{bmatrix}.
 \end{equation*}
 Here $i$ is a hyper-parameter such that
 \begin{equation*}
     1 \le i \le r.
 \end{equation*}
   \item $D'$ is formed from $D$ and $\Delta D$
   \small 
   \begin{equation*}
       D' = D + \Delta D =\begin{bmatrix}
                        \sigma_{1} & \hdots & \hdots & \hdots & \hdots & 0\\
                        0 &  \ddots & \hdots & \hdots & \hdots & \vdots\\
                        \vdots & \hdots & \sigma_{1} & \hdots & \hdots & \vdots\\
                        \vdots & \hdots & \hdots & \sigma_{i+1} & \hdots & \vdots\\
                        \vdots & \hdots & \hdots & \hdots & \ddots & \vdots\\
                        0 & \hdots & \hdots & \hdots & \hdots & \sigma_{r} \\
                       
  \end{bmatrix}. 
   \end{equation*}
  \item $W$ turns to $W'=W +\Delta W$ such that
  \begin{equation*}
     W'= U \cdot \begin{bmatrix}
D & 0 \\
0 & 0 
\end{bmatrix} \cdot V^{T} + U \cdot \begin{bmatrix}
{\Delta D} & 0 \\
0 & 0 
\end{bmatrix} \cdot V^{T},
 \end{equation*}
 where $\Delta W =  U \cdot \begin{bmatrix}
{\Delta D} & 0 \\
0 & 0 
\end{bmatrix} \cdot V^{T}.$

\item To maintain Lipschitz continuity, spectral normalization is applied. Spectral regularized weights $(W_{SR}(W))$ are obtained as follows
\begin{equation*}
    W_{SR}(W) = W'/\sigma(W) = {(W + \Delta W)}/\sigma(W).
\end{equation*}

 \end{enumerate}
%%\sout{\textcolor{blue}{WHAT IS SPECTRAL REGULARIZATION}}
\begin{figure*}[h!]
\centering
\includegraphics[scale=0.35]{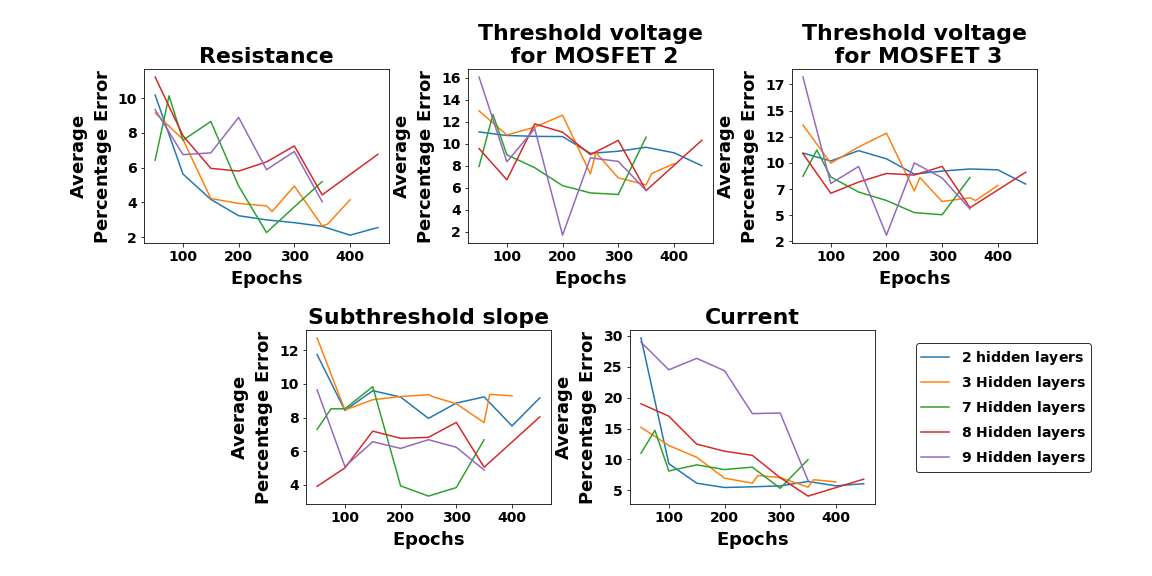}
\caption{\label{figure:2b}Performance of GAN model with different hidden layers w.r.t. Cadence for current reference circuit data.}
\end{figure*}
\begin{figure*}[t!]
     \centering
         
     % \begin{subfigure}[b]{0.24\textwidth}
     %     \centering
     %     \includegraphics[width=\textwidth]{currentref_3l_001_3v3}
     %     %\caption{$Performance\ of\ GAN\ model\ with\ 3\ hidden\\ layers\ learning\ rate\ 0.001\ wrt\ Cadence\ Virtuoso\\ for\ current\ reference\ circuit\ data.$}
     %     \caption{$\alpha=0.001$}
     %     \label{figure:6}
     % \end{subfigure}
    %  \hfill

    %  \hfill
     \begin{subfigure}[b]{0.30\textwidth}
         \centering
         \includegraphics[width=\textwidth]{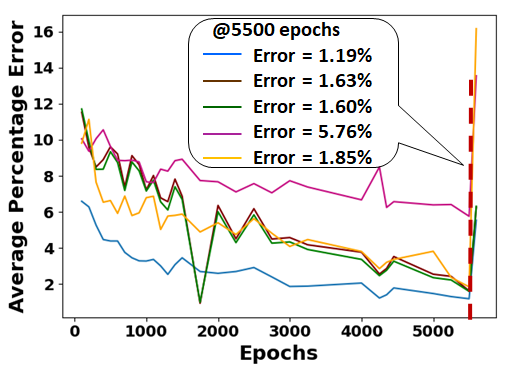}
         %\caption{$Performance\ of\ GAN\ model\ with\ 3\ hidden\\ layers\ learning\ rate\ 0.00025\ wrt\ Cadence\ Virtuoso\\ for\ current\ reference\ circuit\ data.$}
         \caption{$\alpha=0.00025$}
         \label{figure:8}
     \end{subfigure}
    %  \hfill
    % \begin{subfigure}[b]{0.30\textwidth}
    %      \centering
    %      \includegraphics[width=\textwidth,keepaspectratio]{currentref_2l_0005_3v3.png}
    %      %\caption{$Performance\ of\ GAN\ model\ with\ 3\ hidden\\ layers\ learning\ rate\ 0.0005\ wrt\ Cadence\ Virtuoso\\ for\ current\ reference\ circuit\ data.$}
    %      \caption{$\alpha=0.0005$}
    %      \label{figure:9}
    %  \end{subfigure}
       \begin{subfigure}[b]{0.30\textwidth}
         \centering
         \includegraphics[width=\textwidth]{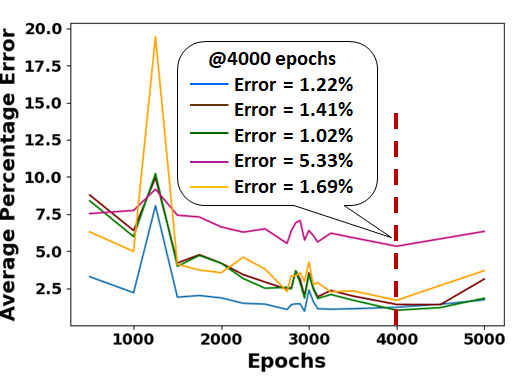}
         %\caption{$Performance\ of\ GAN\ model\ with\ 3\ hidden\\ layers\ learning\ rate\ 0.0005\ wrt\ Cadence\ Virtuoso\\ for\ current\ reference\ circuit\ data.$}
         \caption{$\alpha=0.0005$}
         \label{figure:7}
     \end{subfigure}
     \begin{subfigure}[b]{0.30\textwidth}
         \centering
         \includegraphics[width=\textwidth]{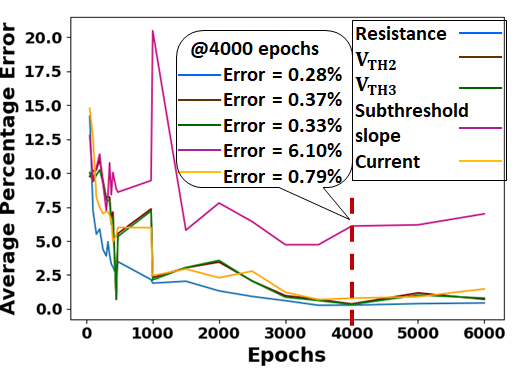}
         %\caption{$Performance\ of\ GAN\ model\ with\ 3\ hidden\\ layers\ learning\ rate\ 0.001\ wrt\ Cadence\ Virtuoso\\ for\ current\ reference\ circuit\ data.$}
         \caption{$\alpha=0.001$}
         \label{figure:6}
     \end{subfigure}
    %  \hfill
    \caption{Here, $\alpha$ represents the learning rate for ADAM optimizer. Performance of GAN model with three layers for Current reference dataset (see Table \ref{table:data}) for different learning rates.}
        \label{figure:58}
\end{figure*}

%%\sout{\textcolor{blue}{ WHY HAVE WE USED THEM}}
We compared the performance of our GAN with spectral normalization and spectral regularization. To obtain quality data augmentation we use density distribution plots and calculate KL divergence between the distributions of generated data and training data to understand and compare the performance. Here, KL divergence is a measure of how one probability distribution differs from another. We found spectral regularization indeed delivers better results. The results are summarized in the next section.
%%\sout{\textcolor{red}{Formulate spectral methods using math notations as done in original paper!}}
\section{Numerical Experiments}
\subsection{Setup of Experiments}
%%\sout{\textcolor{red}{Describe GAN architecture (number of layers, activation functions, conv layers, etc) in detail for both generator and discriminator,}}\textcolor{red}{codes/libraries used.}
We use {\tt Python-3.8.16} and Google Colab for the training of GAN models. Moreover, our implementation uses {\tt Keras-2.9.0} and {\tt Tensorflow-2.9.2}.

% \begin{figure*}[h!]
% \centering
% \includegraphics[scale=0.35]{Current_ref_layersV3.png}
% \caption{\label{figure:2b}Performance of GAN model with different hidden layers w.r.t. Cadence for current reference circuit data.}
% \end{figure*}

\subsubsection{GAN Architecture}
For each dataset, a GAN model consisting of a discriminator model and a
generator model was formed. The discriminator and generator models are neural networks with the same number of layers. The generator network has input dimensions the same as the latent dimensions. Input to the generator is randomly generated latent points. Generator network has output dimensions the same as the dimensions of real data points. Discriminator network has input dimensions the same as the dimensions of
real data points. Discriminator network has output dimension of 1 for the output as classification into artificial or real. For the hidden layers, we have used the leaky ReLU activation function. The output layer for the generator model uses a hyperbolic tangent activation function, whereas, the output layer for the discriminator model uses sigmoid activation. Both these networks in sequential form create the GAN architecture.

\subsubsection{Performance Evaluation of GAN}
%%\sout{\textcolor{blue}{performance evaluation of GAN is 
%open research problem}}
%%\sout{\textcolor{blue}{, currently used methods but how they are good for images only}}

%%\sout{\textcolor{blue}{,how we evaluated the performance}}

Assessing the performance of different GAN models is necessary to compare the quality of synthetic data generated. Numerous quantitative and qualitative measures have been suggested relating to the evaluation and interpretation of generative models \cite{b10}, for example, inception score, Frechet inception distance, average log-likelihood, Parzen window estimates, and visual fidelity. These measures have been proposed to focus on image data generation; moreover, a single standard cannot cover all facets of image generation. Thus, there is no concurrence regarding the best measure \cite{b10}. 
Theis et al. \cite{b11} also pointed out that a model may perform well concerning a measure and not perform well concerning others. They \cite{b11} also suggested that evaluation for generative models needs to be done directly in the context of the intended application.

Thus keeping the above points in mind, the performance analysis of the GAN models was first done with respect to a trained Artificial Neural Network(ANN) model. GAN was used to generate data comprising input features and output features for ANN, then gave the generated input features (generated by GAN generator) to ANN as input. The ANN predicted the value of the output feature for these input feature values. Then both the output values were compared. Despite the small overall percentage error, the generated synthetic datasets had a high percentage error for many individual samples. To get a proper idea of the artificial dataset quality, we changed our approach for evaluation.

We finally evaluated the GAN models for circuit data directly with respect to the simulator, which was the source of our data itself. We select a few features from the generated synthetic dataset and
give them as input to the simulator. Let us call these features as input features and the remaining features as output features. Now we compare the output feature
values from the simulator and the generator. Our metric for comparison
is the average percentage error. Figure \ref{figure:56} shows the complete evaluation process for artificially generated data.

Very few real data samples are used for our experiments to train the GAN. Then artificial data samples are generated using only the generator. Next, we evaluate the GAN performance using these artificial samples, as shown in Figure \ref{figure:56}. If we observe a reduction in error with increasing epochs, we keep training the GAN for more epochs and then save the model with satisfactory performance. If the error does not reduce or keeps increasing, we tune the hyperparameters and try to obtain satisfactory performance. Figure \ref{figure:57} explains the complete experimental process for training the GAN on circuit data.

% \begin{figure}[h]
% \centering
% \includegraphics[scale=0.3]{real_datagen.pdf}
% \caption{\label{figure:55}Dataset generation using simulator.}
% \end{figure}

\subsection{Results}
%%\sout{\textcolor{red}{Add tables, figures, classification accuracy, discuss
%with results how reducing the architectue led to more accuracy, but in some
%cases led to mode collapse. Show with evidence of distribution plots that
%there was less diverse data being generated. Later in the following section
%you show with evidence what you did to resolve.}}

% GAN was used to generate data comprising input features and output features for ANN. The generated input features(generated by GAN generator) to ANN as input. ANN predicted the value of the output feature for these input feature values. Then both the output values were compared. The artificial data generated using generator network was then given to ANN and the error was calculated for the predicted output as shown in Table \ref{table:3}.
% Despite the small overall percentage error, the generated synthetic datasets had a high percentage error for many individual samples. To get a proper idea of the artificial dataset quality, we changed our approach for evaluation.
We evaluated our GAN with respect to a pre-trained ANN model which was trained on the current reference circuit data. 
The performance specifications of this ANN model are listed in Table \ref{table:2}.
\begin{table}
\centering 
\begin{tabular}{|p{4cm}|p{2.5cm}|}
\hline 
     Mean squared error & 1.92e-05\\ \hline 
     Root mean squared error & 0.004\\ \hline
     Mean absolute error & 0.003\\ \hline
     R2 Score & 0.991 \\ \hline
     Mean Percentage error & 0.88 \\ \hline
     %1.919809945420651e-05  & 1.919809945420651e-05 &  0.003309683092751534 & 0.9999808479331461 & 0.8783166051796849\\ \hline 

\end{tabular}
\caption{\label{table:2}Performance of trained ANN w.r.t. Cadence. %%\sout{\textcolor{red}{keep it to two digits after decimal}}
}
\end{table}

\begin{table}
\centering
    \begin{tabular}{|p{0.8cm}|p{1.2cm}|p{1.2cm}|p{1.2cm}|p{1.2cm}|}
    \hline
      \textbf{ } & \textbf{Model 1} & \textbf{Model 2} & \textbf{Model 3} & \textbf{Model 4} \\ \hline
      MSE & 4.6e-13 & 3.7e-13 &	1.2e-13 & 3.2e-13\\ % <--
      RMSE & 6.8e-07	& 6.1e-07 & 3.5e-07 & 5.6e-07\\ %<--
      MAE & 5.9e-07 &	4.8e-07 &	2.6e-07 &	4.3e-07\\\hline % <--
    \end{tabular}
    \caption{\label{table:3}Performance of GAN w.r.t. ANN for current reference circuit data. Here MSE stands for mean square error, RMSE stands for root mean square error, and MAE stands for mean average error.}
\end{table}
The artificial data generated using generator network was then given to ANN and the error was calculated for the predicted output as shown in Table \ref{table:3}.
However, this evaluation method had drawbacks. Despite the small mean squared error, we observed that the synthetic data-sets generated had high absolute percentage error for many individual samples.

\begin{figure*}
     \centering
         
     \begin{subfigure}[b]{0.445\textwidth}
         \centering
         \includegraphics[width=0.75\textwidth]{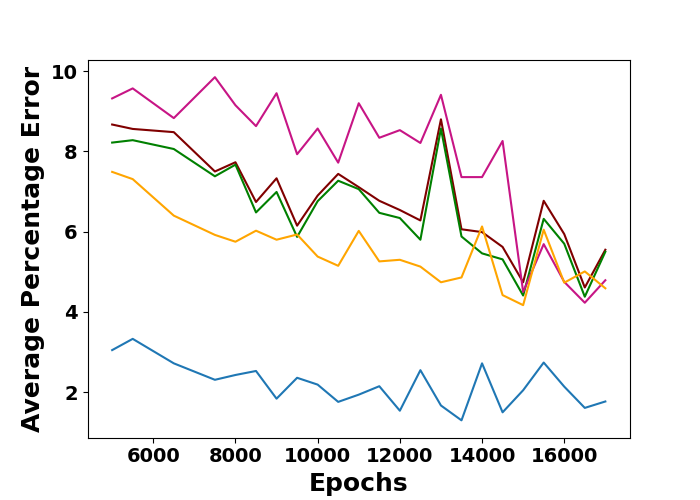}
         \caption{Performance of GAN model with 3 hidden layers and learning rate 0.0005 with spectral regularization w.r.t. Cadence Virtuoso\ for current reference circuit data.}
         \label{figure:11}
     \end{subfigure}
    %  \hfill
     \begin{subfigure}[b]{0.38\textwidth}
         \centering
         \includegraphics[width=\textwidth]{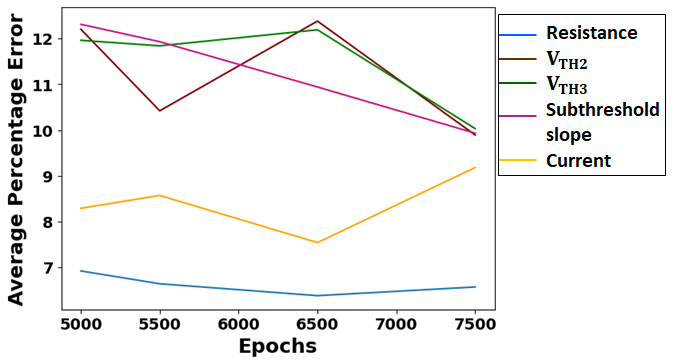}
         \caption{Performance of GAN model with 3 hidden layers learning rate 0.0005 and spectral normalization w.r.t. Cadence Virtuoso for current reference circuit data.}
         \label{figure:10}
     \end{subfigure}
    %  \hfill
        \caption{Performance of GAN model with 3 layers for Current reference dataset after applying spectral methods.}
        \label{figure:59}
\end{figure*}

Therefore, we compared the output feature values from the generator and the simulator to get the right idea of performance (see Figures \ref{figure:58},\ref{figure:2b},\ref{figure:59}). 

\subsubsection{Hyperparameter Search} We began our hyper-parameter search by looking for the optimum number of layers. We observed the performance of the GAN model with varying numbers of hidden layers (refer to Figure~\ref{figure:2b}). 
% We observed the performance of different models with different numbers of layers.
% % Table \ref{table:4} shows the effect of increasing epochs on a complex GAN architecture with seven hidden layers.
% Figure~\ref{figure:2} shows the effect of increasing epochs on average percentage error for a complex GAN architecture with seven hidden layers.
% % Table \ref{table:5} shows the effect of increasing epochs on a complex GAN architecture with eight hidden layers.
% Figure~\ref{figure:3} shows the effect of increasing epochs on average percentage error for a complex GAN architecture with eight hidden layers.
% % Table \ref{table:6} shows the effect of increasing epochs on a complex GAN architecture with nine hidden layers.
% Figure~\ref{figure:4} shows the effect of increasing epochs on average percentage error for a complex GAN architecture with nine hidden layers. 

We observed that the increasing number of hidden layers seemed too complex for our data sets, and this benefited very little in decreasing the percentage error. Moreover, more layers will lead to more computation time for calculating spectral values while using spectral regularization or normalization. Hence, we reduced the number of layers.
From Figure~\ref{figure:2b}, we find that there is no clear choice for the number of hidden layers as two and three hidden layers both perform well for all input features simultaneously. We used three hidden layers for the current reference dataset as a tradeoff between model complexity and performance. For some datasets, we used two hidden layers with comparable performance.
We next experiment with different learning rates. Figure~\ref{figure:58} shows that a learning rate of around 0.0005 provides a low percentage error.

% On observing %Table \ref{table:9}, 
% Figure~\ref{figure:7} and %Table \ref{table:10}, 
% Figure~\ref{figure:8} we find the percentage error reduced to a very low value.
On further investigation, we find that the GAN suffers from mode collapse, i.e., the model becomes good at generating only a few points from the entire training data distribution. Thus, the generator generates only these points and successfully fools the discriminator. This is evident in Figure \ref{figure:12}, where the density distribution of generated data without spectral regularization or normalization has peaked in some areas of training data only. This shows the generated data is less diverse. Also, the KL divergence is high for the generated data distribution (refer Figure \ref{figure:12}).

\begin{table*}[h!]
\centering
    \begin{tabular}{|p{3.2cm}|p{1.6cm}|p{1.8cm}|p{2.3cm}|p{3cm}|p{3.1cm}|}
    \hline
      \textbf{Complex digital circuit} & \textbf{Simulated delay(ps)} & \textbf{Predicted delay with real data(ps) } & \textbf{Predicted delay with real + artificial data(ps)} & \textbf{Error in predicted delay with real data(\%) } & \textbf{Error in predicted delay with real + artificial data(\%) } \\ \hline
      ISCAS C17 & 10.7 & 11 &	10.6 & 3.08 & 0.65\\ \hline % <--
      4-bit ripple carry adder & 41	& 43.1 & 42 & 5 & 2.4\\ \hline %<--
    \end{tabular}
    \caption{\label{table:complexcircuits}Comparison of percentage error w.r.t. simulated delay for the predicted delay in complex digital circuits.}
\end{table*}

%%\sout{\textcolor{blue}{Points-Current reference wrt ann, gbr, how high 
%\% error for some samples and low overall error, Comparison with simulator 'cadence', effect of architecture, learning rate, mode collapse in my words}}

\begin{table*}[h!]
\centering 
% \begin{tabular}{|c|c|c|}
% \begin{tabular}{|p{4.8cm}|p{2.67cm}|p{2.67cm}|p{2.67cm}|}
% \begin{tabular}{|c|c|c|c|c|c|c|c|c|}
\begin{tabular}{|p{4.8cm}|p{1.05cm}|p{1.05cm}|p{1.05cm}|p{1.05cm}|p{1.05cm}|p{1.05cm}|p{1.05cm}|p{1.05cm}|}
\hline 
    Delay dataset  & \multicolumn{8}{|c|}{Average percentage error} \\ %\hline
    \cline{2-9}
    {} & delay lh node a & delay hl node a & delay lh node b & delay hl node b & 
 delay lh node c & delay hl node c & delay lh node d & delay hl node d\\ \hline  
    NOT gate   & 7.00 & 5.81 & - & - & - & - & - & - \\ \hline
    Two input NAND gate    & 5.43 & 6.01 & 6.06 & 6.52 & - & - & - & -  \\ \hline 
    Two input AND gate     & 5.75 & 5.26 & 6.42 & 5.81 & - & - & - & -  \\ \hline 
    Two input NOR gate     & 7.30 & 4.87 & 7.34 & 5.36 & - & - & - & - \\ \hline
    Two input OR gate     & 6.05 & 5.77 & 5.49 & 6.37 & - & - & - & - \\ \hline
    Two input XOR gate     & 5.83 & 5.90 & 5.25 & 4.63 & - & - & - & - \\ \hline
    Three input AND-OR circuit     & 8.85 & 7.25 & 6.80 & 8.05 & 5.9 & 8.74 & - & - \\ \hline
    Full adder     & 4.50 & 4.54 & 4.24 & 3.57 & 3.80 & 3.30 & - & - \\ \hline
    2:1 Multiplexer     & 5.37 & 4.38 & 5.45 & 4.78 & 5.84 & 5.58 & - & -\\ \hline
    Three input NAND gate     & 9.15 & 4.44 & 8.71 & 4.27 & 8.37 & 6.19 & - & -\\ \hline
    Three input AND gate     & 5.41 & 5.10 & 4.64 & 4.64 & 5.41 & 4.19 & - & -\\ \hline
    Three input NOR gate     & 7.09 & 4.32 & 6.62 & 4.25 & 7.41 & 4.37 & - & -\\ \hline 
    Four input AND-OR circuit (AO22)    & 5.13 & 3.71 & 4.60 & 4.13 & 5.50 & 4.15 & 5.01 & 4.77 \\ \hline
    Four input AND-OR circuit (AO31)    & 5.40 & 4.91 & 5.50 & 6.11 & 5.15 & 5.42 & 5.48 & 6.04 \\ \hline
    
\end{tabular}
\caption{\label{table:errordigitaldata}Percentage error obtained for different digital circuit datasets used in this work.}
\end{table*}

\subsubsection{Effect of Spectral Normalization and Spectral Regularization for Avoiding Mode Collapse}
\begin{figure*}[h]
\centering
\includegraphics[max size={\textwidth}{\textheight}]{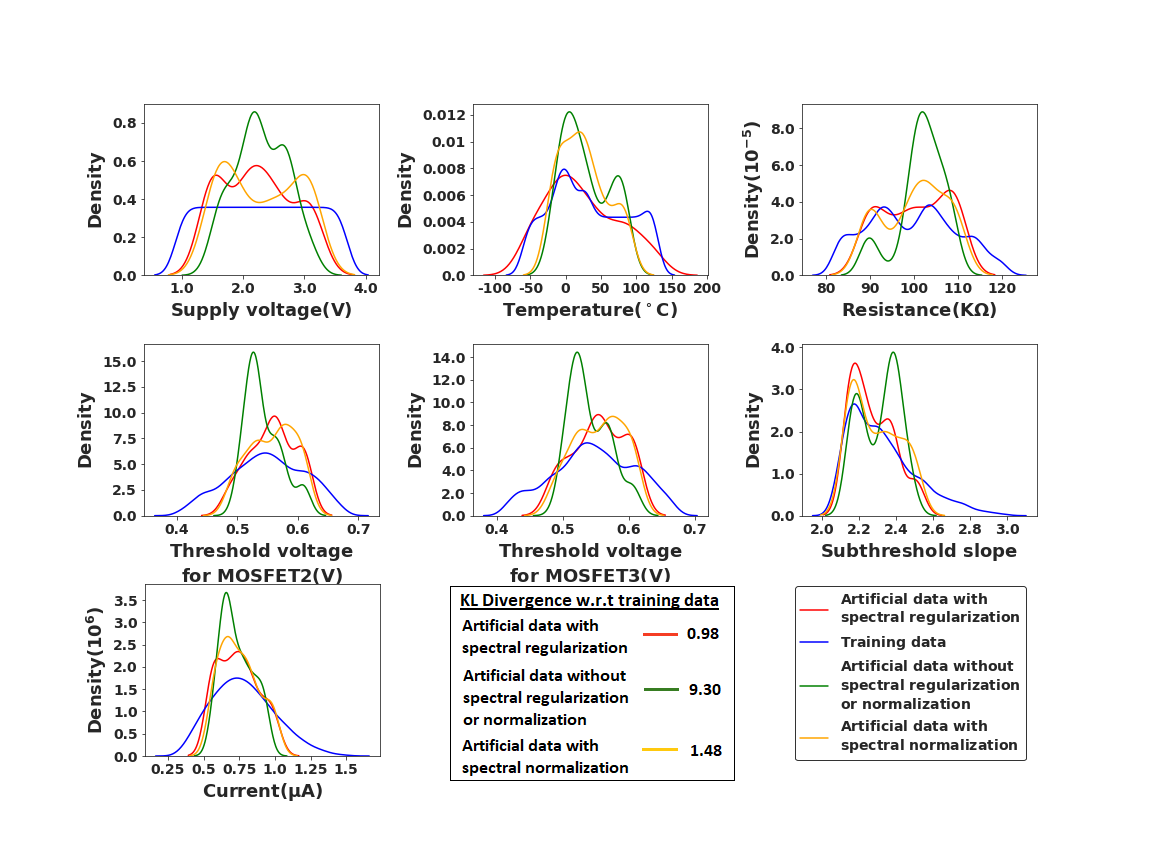}
\caption{\label{figure:12}Comparison of the distribution of training dataset for current reference circuit and various generated datasets. KL divergence between data distributions from different generators with respect to training data is also shown in the lower left. (KL divergence measures how one probability distribution differs from another.)}
\end{figure*}

%\sout{\textcolor{red}{In this section, describe how we dealt with mode collapse using spectral normalization. Also describe whether we lost on diversity of the generated dataset while doing so.}}
%\sout{\textcolor{blue}{Observations in my words}}
To deal with the issue of mode collapse, we experimented with the spectral normalization proposed by Takeru Miyato et al. \cite{b9} and spectral regularization proposed by Kanglin Liu et al. \cite{b8}. We observe that both of these methods successfully avoid mode collapse in our models, as seen in the distribution plot in Figure \ref{figure:12}. The plots show that although the error percentage is slightly more, the generated data has a diverse distribution. The increase in error can be attributed to the absence of mode collapse.

On comparing spectral regularization and spectral normalization, we found spectral regularization to be performing better with respect to distribution plots (Figure \ref{figure:12}) as well as average percentage error, refer to %Table \ref{table:12} and Table \ref{table13}, 
Figure \ref{figure:11} and Figure \ref{figure:10}. Moreover, Figure \ref{figure:12} also confirms that the Kl divergence for the data distribution generated using spectral regularization is lower than the KL divergence for the data distribution generated using spectral normalization.

\begin{figure*}
     \centering
         
     \begin{subfigure}[b]{0.325\textwidth}
         \centering
         \includegraphics[width=\textwidth]{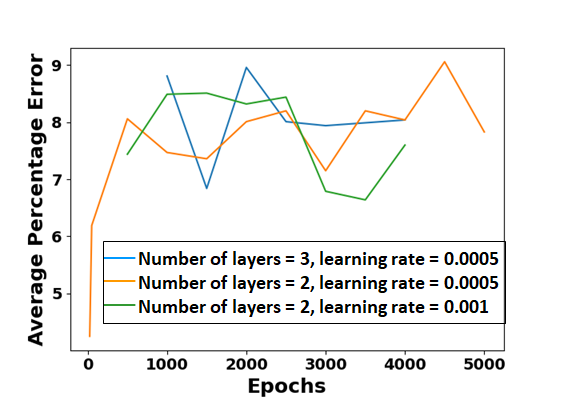}
         \caption{$GAN\ models\  wrt\ Microcap\ for\ LDO\\ circuit.\ (Best\ performance\ with\ 2\ hidden\\ layers\ and\ learning\ rate\ of\ 0.001)$}
         \label{figure:15}
     \end{subfigure}
    %  \hfill
     \begin{subfigure}[b]{0.325\textwidth}
         \centering
         \includegraphics[width=\textwidth]{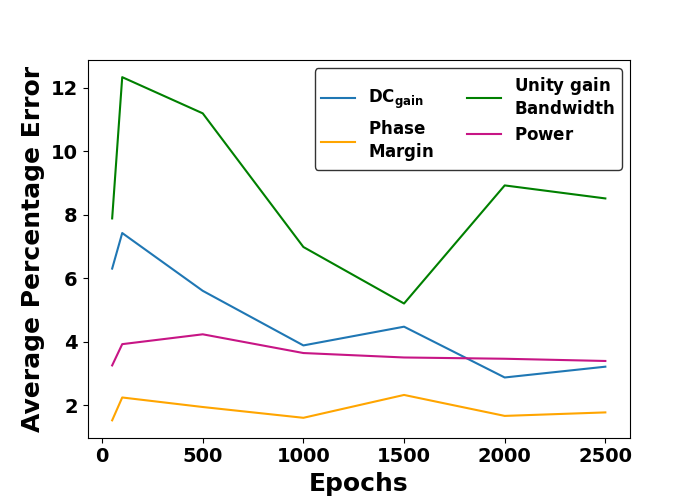}
         
         \caption{$GAN\ model\ with\ 3\ hidden\ layers\ and\\ learning\ rate\ of\ 0.0005\ wrt\ Cadence\\ Virtuoso\ for\ OTA\ circuit.$}
         \label{figure:16}
     \end{subfigure}
    %  \hfill
     %     \begin{subfigure}[b]{0.43\textwidth}
     %     \centering
     %     \includegraphics[width=\textwidth]{SR_OTA_2l_0005_3v3.png}
     %     \caption{$Performance\ of\ GAN\ model\ with\ 2\ hidden\ layers\\ learning\ rate\ 0.0005\ and\ spectral\ regularization\ wrt\\ Cadence\ Virtuoso\ for\ OTA\ circuit\ data$}
     %     \label{figure:17}
     % \end{subfigure}
      \begin{subfigure}[b]{0.325\textwidth}
         \centering
         \includegraphics[width=\textwidth]{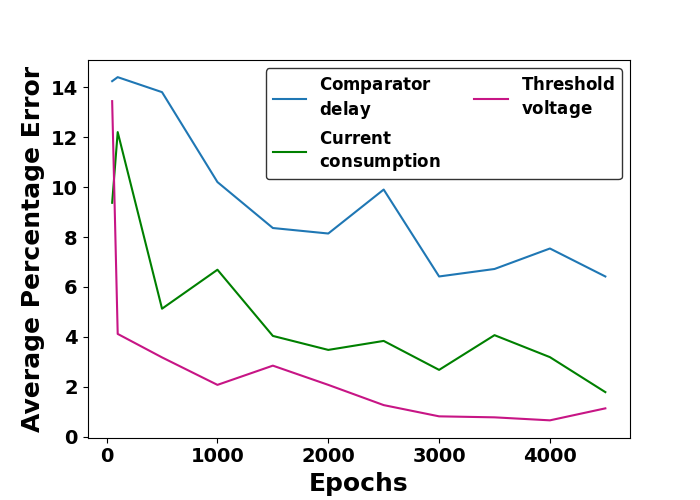}
         \caption{$GAN\ model\ with\ 3\ hidden\ layers\ and\\ learning\ rate\ of\ 0.0005\ wrt\ Cadence\\ Virtuoso\ for\ Comparator\ circuit.$}
         \label{figure:18}
     \end{subfigure}
     
        \caption{Performance of GAN models with spectral regularization for different analog circuit datasets.}
        \label{figure:60}
\end{figure*}

\begin{figure*}
     \centering
         
     \begin{subfigure}[b]{0.325\textwidth}
         \centering
         \includegraphics[width=\textwidth]{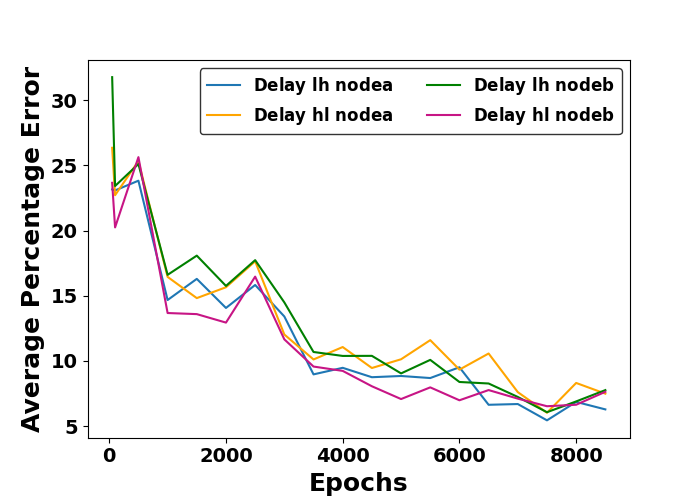}
         \caption{$Model\ with\ 3\ hidden\ layers\ for\\ delay\ in\ two\ input\ NAND\ gate.$}
         \label{figure:31}
     \end{subfigure}
     \begin{subfigure}[b]{0.325\textwidth}
         \centering
         \includegraphics[width=\textwidth]{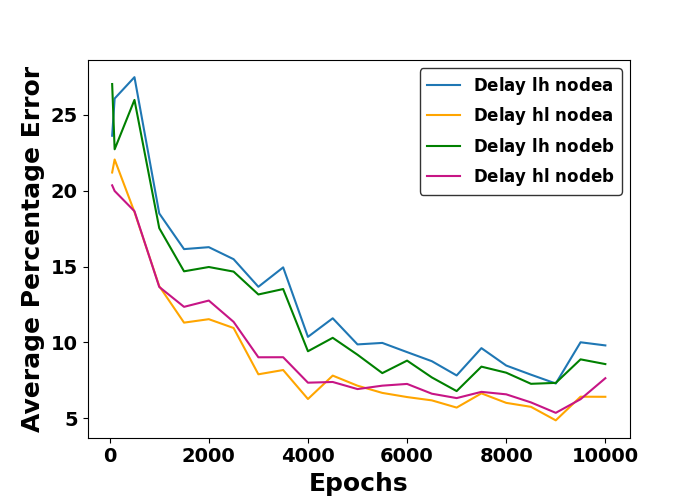}
         \caption{$Model\ with\ 2\ hidden\ layers\ for\\ delay\ in\ two\ input\ NOR\ gate.$}
         \label{figure:33}
     \end{subfigure}
      \begin{subfigure}[b]{0.325\textwidth}
         \centering
         \includegraphics[width=\textwidth]{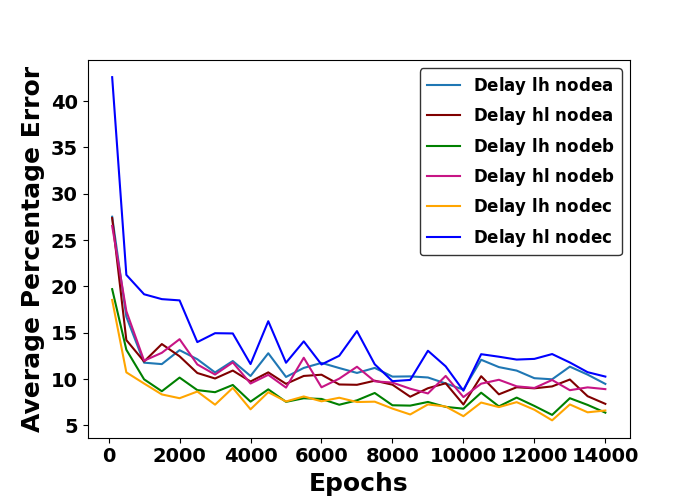}
         \caption{$Model\ with\ 3\ hidden\ layers\ for\ delay\\ in\ three\ input\ AND-OR\ (AO12)\ gate.$}
         \label{figure:42}
    \end{subfigure}
    \vspace{-0.16cm}
     \begin{subfigure}[b]{0.325\textwidth}
         \centering
         \includegraphics[width=\textwidth]{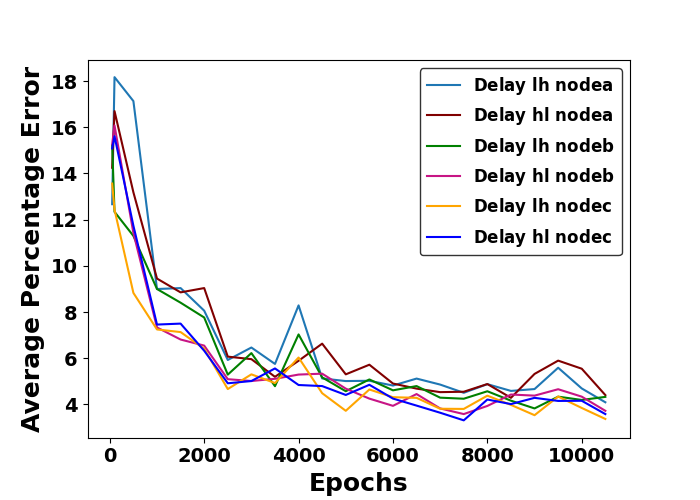}
         \caption{$Model\ with\ 3\ hidden\ layers\ for\\ delay\ in\ Full\ adder.$}
         \label{figure:43}
     \end{subfigure}
     \begin{subfigure}[b]{0.325\textwidth}
         \centering
         \includegraphics[width=\textwidth]{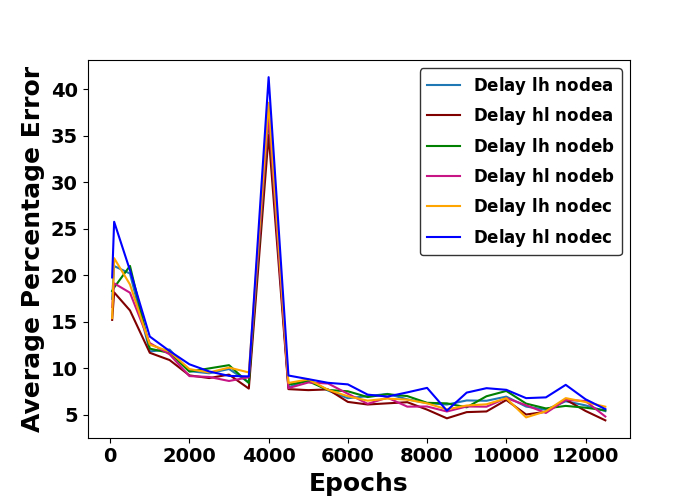}
         \caption{$Model\ with\ 3\ hidden\ layers\ for\\ delay\ in\ $2:1$\ Multiplexer.$}
         \label{figure:44}
     \end{subfigure}    
    \begin{subfigure}[b]{0.325\textwidth}
         \centering
         \includegraphics[width=\textwidth]{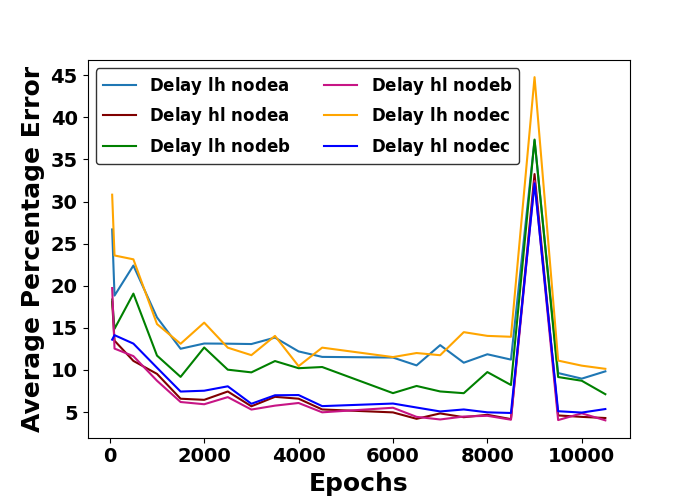}
         \caption{$Model\ with\ 3\ hidden\ layers\ for\ delay\\ in\ three\ input\ NAND\ gate.$}
         \label{figure:45}
     \end{subfigure}  
     \begin{subfigure}[b]{0.325\textwidth}
         \centering
         \includegraphics[width=\textwidth]{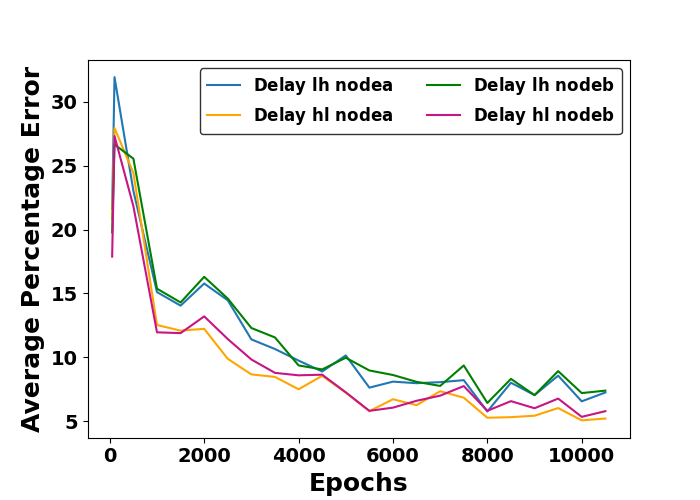}
         \caption{$Model\ with\ 3\ hidden\ layers\ for\\ delay\ in\ two\ input\ AND\ gate.$}
         \label{figure:29}
     \end{subfigure}    
     \begin{subfigure}[b]{0.325\textwidth}
         \centering
         \includegraphics[width=\textwidth]{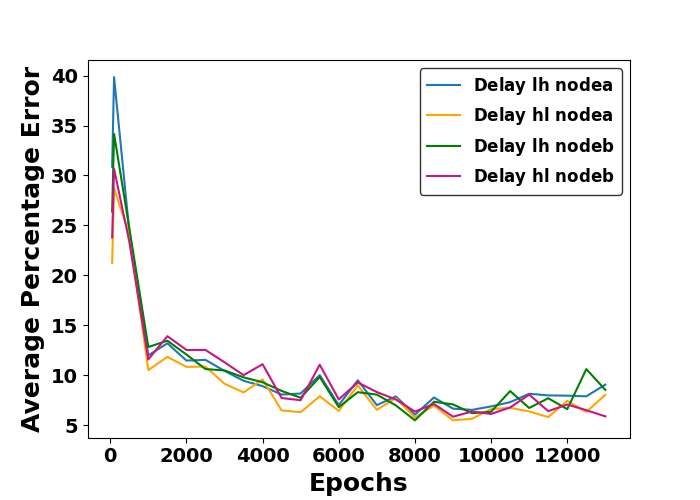}
         \caption{$Model\ with\ 3\ hidden\ layers\ for\\ delay\ in\ two\ input\ OR\ gate.$}
         \label{figure:48}
     \end{subfigure}
    %  \hfill
     \begin{subfigure}[b]{0.325\textwidth}
         \centering
         \includegraphics[width=\textwidth]{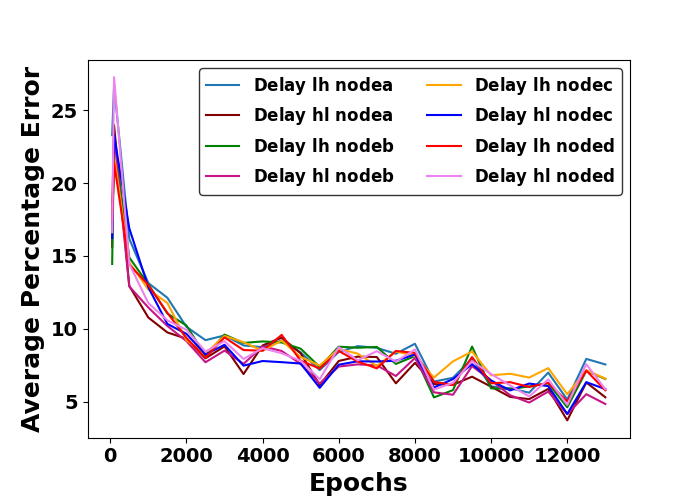}
         \caption{$Model\ with\ 2\ hidden\ layers\ for\ delay\\ in\ four\ input\ AND-OR22\ gate.$}
         \label{figure:49}
     \end{subfigure}
     \begin{subfigure}[b]{0.325\textwidth}
         \centering
         \includegraphics[width=\textwidth]{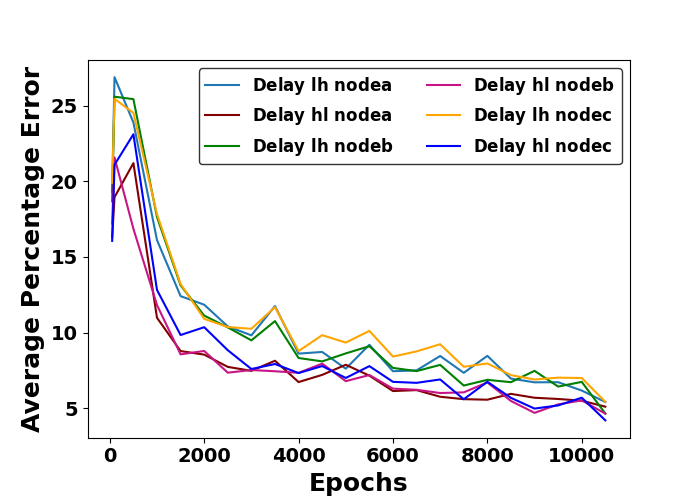}
         \caption{$Model\ with\ 4\ hidden\ layers\ for\\ delay\ in\ three\ input\ AND\ gate.$}
         \label{figure:50}
     \end{subfigure}
     \begin{subfigure}[b]{0.325\textwidth}
         \centering
         \includegraphics[width=\textwidth]{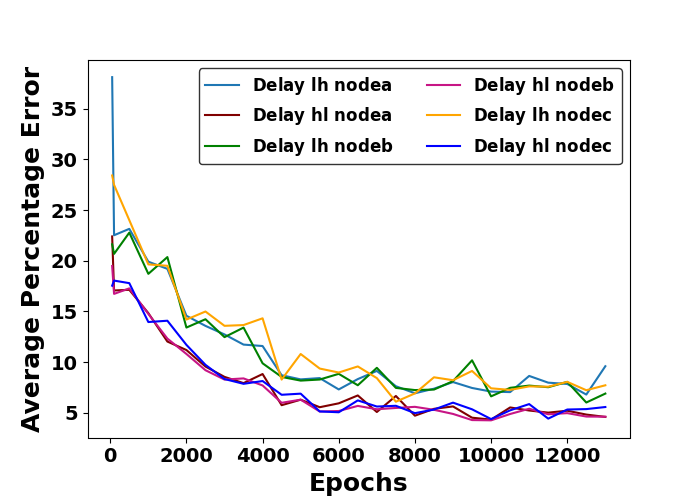}
         \caption{$Model\ with\ 4\ hidden\ layers\ for\\ delay\ in\ three\ input\ NOR\ gate.$}
         \label{figure:51}
     \end{subfigure}
    %  \hfill
    \begin{subfigure}[b]{0.325\textwidth}
         \centering
         \includegraphics[width=\textwidth]{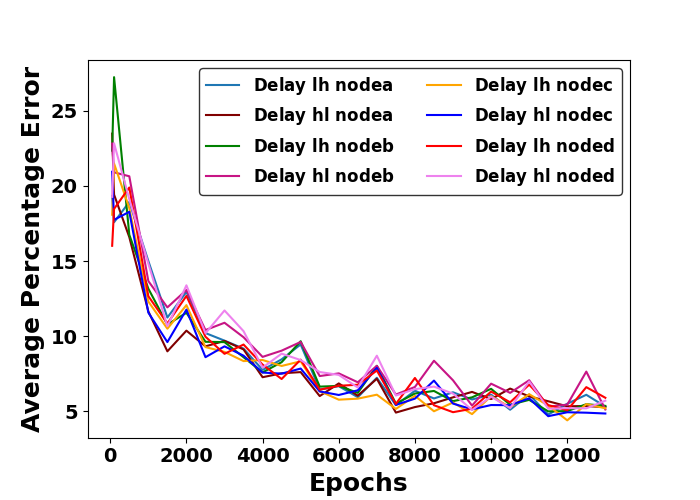}
         \caption{$Model\ with\ 4\ hidden\ layers\ for\ delay\\ in\ four\ input\ AND-OR31\ gate.$}
         \label{figure:52}
     \end{subfigure}
        \caption{Performance of GAN models with learning rate=0.0005 and  spectral regularization, for delay datasets of different digital circuits wrt HSPICE.}
        \label{figure:61}
\end{figure*}

\begin{figure*}[h]
\centering
\includegraphics[max size={\textwidth}{\textheight}]{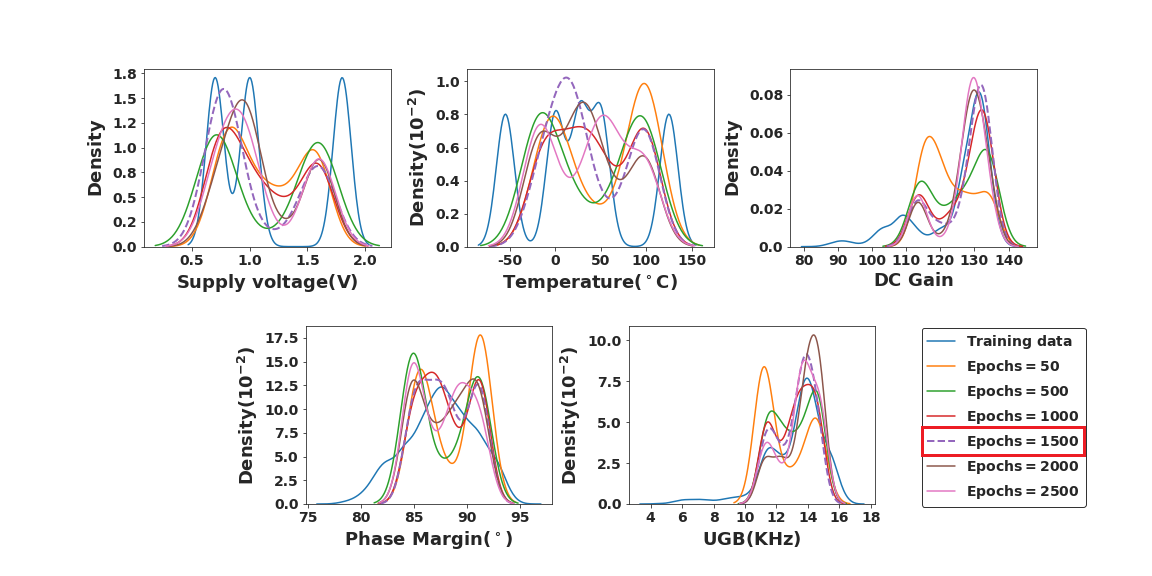}
\caption{\label{figure:23}Comparison of distribution of training dataset for OTA circuit and various generated datasets using a generator with three hidden layers and learning rate 0.0005 at different epochs. (Epoch with lowest percentage error highlighted in red.)}
\end{figure*}

\begin{figure*}[h]
\centering
\includegraphics[max size={\textwidth}{\textheight}]{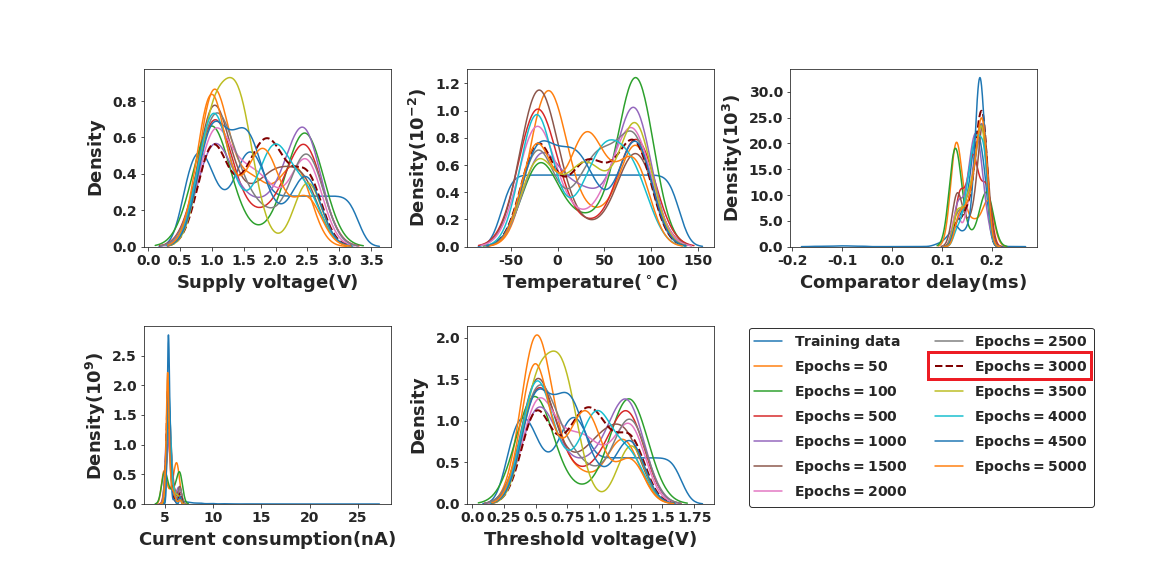}
\caption{\label{figure:25}Comparison of distribution of training dataset for comparator circuit and various generated datasets using a generator with three hidden layers and learning rate 0.0005 at different epochs. (Epoch with lowest percentage error highlighted in red.)}
\end{figure*}

\begin{figure*}[h]
\centering
\includegraphics[max size={\textwidth}{\textheight}]{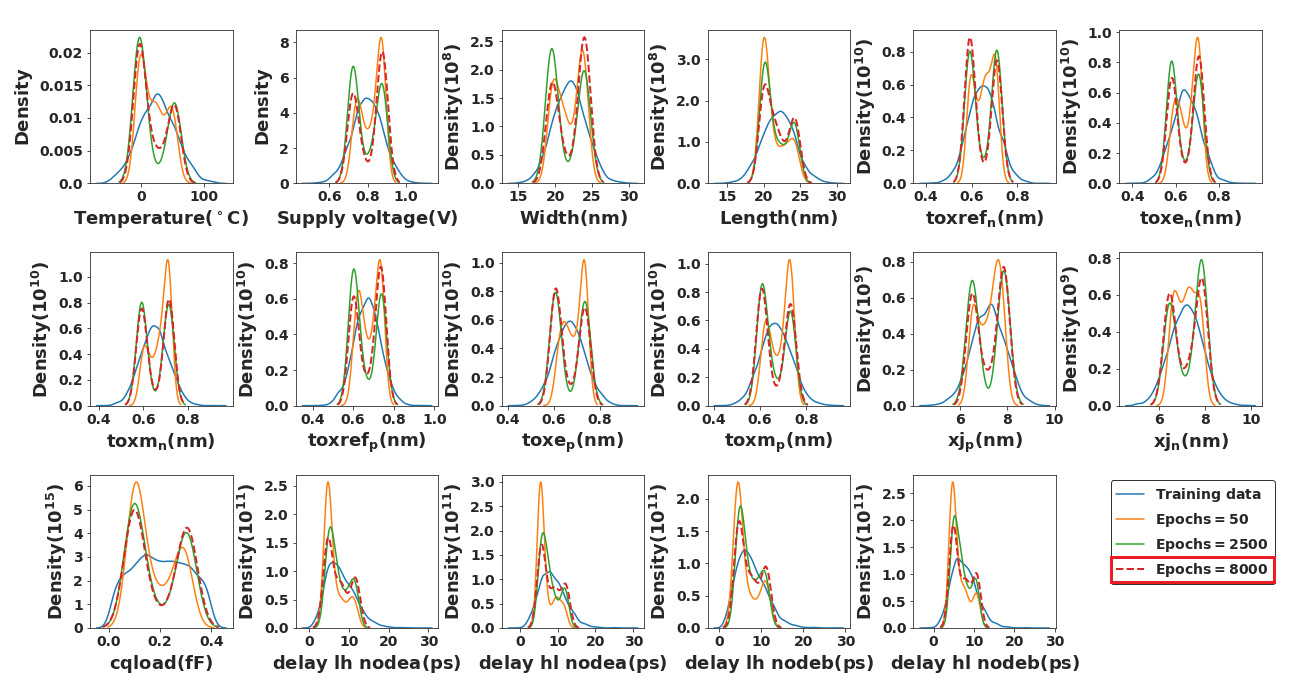}
\caption{\label{figure:26}Comparison of distribution of training dataset for delay in two input AND gate circuit and various generated datasets using a generator with three hidden layers and learning rate 0.0005 at different epochs. (Epoch with lowest percentage error highlighted in red.)}
\end{figure*}

\begin{figure*}[h]
\centering
\includegraphics[max size={\textwidth}{\textheight}]{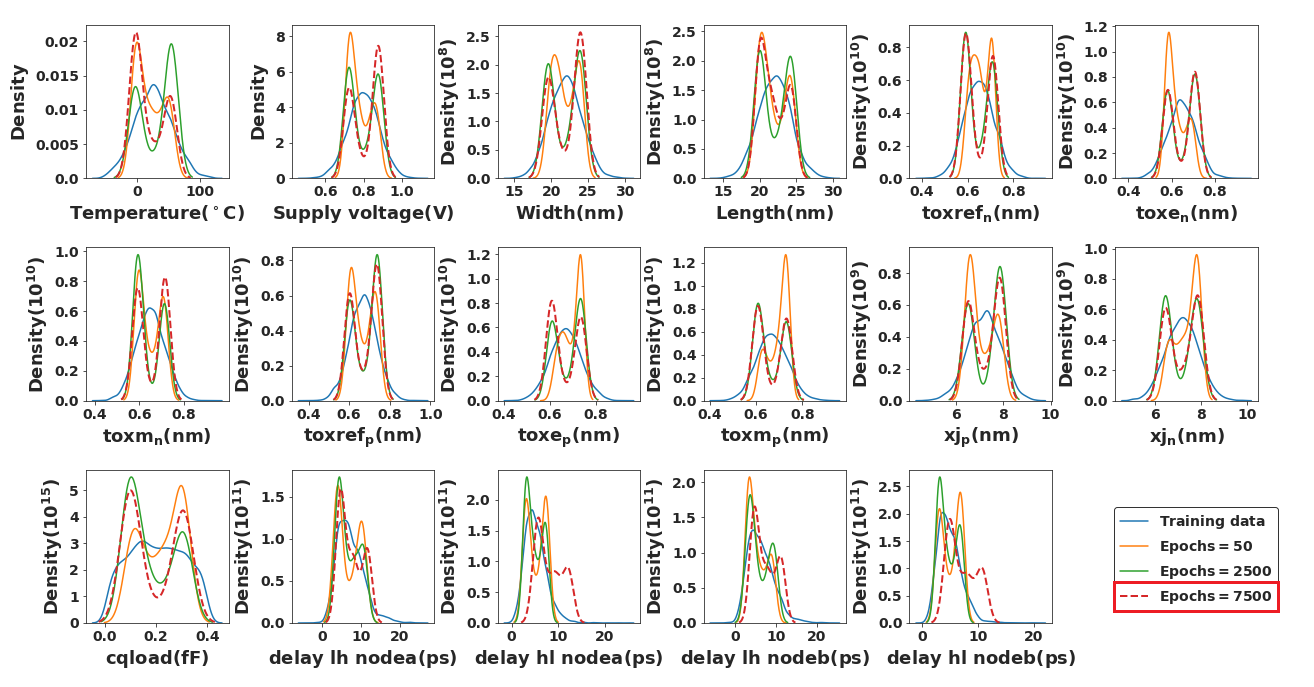}
\caption{\label{figure:27}Comparison of distribution of training dataset for delay in NAND gate circuit and various generated datasets using a generator with three hidden layers and learning rate 0.0005 at different epochs. (Epoch with lowest percentage error highlighted in red.)}
\end{figure*}

We train the GAN model with spectral regularization on various other similar datasets from different analog circuits (Refer Table \ref{table:data}) and digital circuits (Refer Table \ref{table:digitaldata}). Figure \ref{figure:60} shows a reduction in error with increasing epochs for data from other analog circuits, Figure \ref{figure:61} %{and figure \ref{figure:65}}
shows the same for delay datasets from various digital circuits. Table \ref{table:errordigitaldata} shows the low percentage error obtained for various digital circuits. Figure %{\ref{figure:22},}
\ref{figure:23},\ref{figure:25},\ref{figure:26} and \ref{figure:27} show the distribution density plots fitting towards the training data distribution with increasing epochs for few circuits. These figures also ensure that the GAN is free from mode collapse. Hence, we can infer that the generated artificial data is high quality and can be applied to other applications.

\subsubsection{Experiments on Complex Circuits}
As proposed earlier, data augmentation and synthetic data generation can prove helpful for training good ML models for different electronic designs. We experimented with training ML models for some complex digital circuits. We compared the performance of the same model by training with very few real data samples and then training after adding the synthetic data to increase the training data size. We find that the model performs better when we add artificial data samples. The above digital circuits(Refer Table \ref{table:digitaldata}) are the building blocks for various complex digital circuits. Thus, we used the data generated for these basic circuits for training ML models for complex digital circuits.

We used ISCAS benchmark C17 circuit and a 4-bit ripple carry adder for our experiments. We tested a gradient-boosting regression model to predict the delays for the complex circuits. After training on data, the model predicted the delay for individual digital blocks/gates in the circuit, which we used to find the overall delay in the circuit. As shown in Table \ref{table:complexcircuits}, we find a decrease in percentage error for predicted delay when we use the artificial data in addition to actual data for training the ML model for the circuits. Table \ref{table:complexcircuits} shows the decrease is more than 50\% of the original error.

% \begin{figure*}[h]
% \centering
% \includegraphics[max size={\textwidth}{\textheight}]{OTACv3.png}
% \caption{\label{figure:24}Comparison of distribution of training dataset for OTA circuit and various generated datasets using a generator with two hidden layers and learning rate 0.0005 at different epochs .}
% \end{figure*}

% \begin{figure*}[h]
% \centering
% %\includegraphics[max size={\textwidth}{\textheight}]{NOR2C_delay_distributionv3.png}
% \includegraphics[max size={\textwidth}{\textheight}]{NOR2C_delay_distributionv4.png}
% \caption{\label{figure:28}Comparison of distribution of training dataset for delay in NOR gate circuit and various generated datasets using a generator with three hidden layers and learning rate 0.0005 at different epochs.}
% \end{figure*}

% \begin{figure*}[h]
% \centering
% \includegraphics[max size={\textwidth}{\textheight}]{AO12J_delay_distribution.png}
% \caption{\label{figure:28}Comparison of distribution of training dataset for delay in NOR gate circuit and various generated datasets using a generator with three  wide hidden layers and learning rate 0.0005 at different epochs .}
% \end{figure*}

\section{Conclusion}
This paper presents an artificial data generation method for circuits to aid the training of ML models for design automation, tuning, optimization, etc. Model accuracy is heavily dependent on the quantity and quality of training data, but large amounts of training data for electronic circuits can be computationally expensive or practically difficult to obtain. The generated synthetic data is beneficial for training the models when the training data is scarce. 
% GANs, which have been used earlier for image and audio data and have provided promising results in terms of quality and speed, have been used to create artificial data for various electronic circuits. 
GANs have been used in image and audio data modality and have provided promising results in terms of quality and speed. This work adapts GAN to create artificial data for various electronic circuits. 
The training data are obtained by various simulations in the Cadence Virtuoso, HSPICE, and Microcap design environment with TSMC 180nm and 22nm CMOS technology nodes. An evaluation methodology using the simulators is proposed to evaluate the quality of generated data. Spectral regularization has been used to avoid mode collapse in GAN. Artificial data has been generated and tested for six analog and fourteen basic digital circuit designs. The experimental results confirm a low average percentage error for the generated data. The proposed artificial data has finally been applied to gradient-boosting regression models for predicting delays in the ISCAS benchmark C17 circuit and a four-bit ripple carry adder. The simulation results show a reduction in model percentage error by more than 50\% of the previous percentage error when additional artificial data is used for training. 
 
The proposed methodology for generating artificial data has the potential to be applied to many other circuit designs and provide greater accuracy to ML models, especially when the training data is scarce and is challenging to obtain. 
% \section{Acknowledgements}
% We thank IHub-Data, IIIT Hyderabad, for supporting this research.  
\clearpage
%% The Appendices part is started with the command \appendix;
%% appendix sections are then done as normal sections

\appendix

%% If you have bibdatabase file and want bibtex to generate the
%% bibitems, please use
%%
 \bibliographystyle{elsarticle-num} 
 \bibliography{cas-refs}

%% else use the following coding to input the bibitems directly in the
%% TeX file.

% \begin{thebibliography}{00}

% %% \bibitem{label}
% %% Text of bibliographic item

% \bibitem{}

% \end{thebibliography}
\end{document}